\documentclass[sigconf,screen,nonacm]{acmart}

\AtBeginDocument{%
  }

\usepackage{booktabs}
\usepackage{subcaption}
\usepackage{amsmath}
\usepackage{algorithm}
\usepackage{algorithmic}
\usepackage{multirow}
\usepackage{xcolor}
\usepackage{graphicx}

\newcommand{\method}{ImplicitTerrain{V2}}
\newcommand{\psis}{\Psi_{\mathrm{s}}}     %
\newcommand{\psig}{\Psi_{\mathrm{g}}}     %

\newcommand{\ie}{\textit{i.e.}}
\newcommand{\eg}{\textit{e.g.}}
\newcommand{\etal}{\textit{et al.}}

\begin{document}

\title{ImplicitTerrain{V2}: Wavelet-Guided Spatially Adaptive Neural Terrain Representation}

\author{Haoan Feng}
\affiliation{%
  \institution{University of Maryland, College Park}
  \state{MD}
  \country{USA}
}
\email{hfengac@umd.edu}

\author{Xin Xu}
\affiliation{%
  \institution{University of Maryland, College Park}
  \state{MD}
  \country{USA}
}
\email{xinxu629@umd.edu}

\author{Leila De Floriani}
\affiliation{%
  \institution{University of Maryland, College Park}
  \state{MD}
  \country{USA}
}
\email{deflo@umd.edu}

\renewcommand{\shortauthors}{Feng et al.}

\begin{abstract}
Digital elevation models (DEMs) underpin terrain analysis in Geographic Information Systems (GIS), but commonly as raster representation, they rely on interpolation for off-grid sampling and finite-difference operators for derivative-based analysis.
Implicit neural representations (INRs) offer a continuous alternative, but prior terrain INRs lack explicit frequency control, neglect the gradient structure of terrain, and remain too large and costly to train for practical deployment.
We present \method{}, which advances terrain INRs toward a compact, efficient neural terrain data format by combining a spectral control mechanism with wavelet-guided spatial adaptivity, derivative-aware supervision, and post-training model compression.
At its core, a \emph{wavelet complexity field} (WCF) derives spatially-adaptive frequency masks from analytically computed wavelet coefficients, localizing high-frequency capacity to complex terrain regions.
The same field guides \emph{complexity-aware adaptive sampling} that concentrates training in high-complexity regions, while \emph{gradient matching} applies extra supervision to enforce the smooth manifold structure of terrain DEMs for improved derivative fidelity.
Post-training mixed-precision quantization and entropy coding reduce storage to 1.23\,bpp with a 0.28\,dB PSNR drop.
On 50 morphologically diverse Swiss terrain tiles, \method{} reaches 66.25\,dB end-to-end PSNR, improving over the prior work by 5.70\,dB while using 3.2$\times$ fewer parameters and training in 55\,s per tile on a single GPU.
Our compressed neural format is competitive with several established DEM codecs in rate-distortion performance, while additionally supporting off-grid point queries, closed-form derivative evaluation, and resolution-independent reconstruction, which may benefit many downstream GIS applications.

\end{abstract}



\maketitle

\let\oldincludegraphics\includegraphics
\renewcommand{\includegraphics}[2][]{%
    \oldincludegraphics[#1]{#2}%
    \vspace{-0.3cm}%
}

\section{Introduction}
\label{sec:intro}

Terrain representation underpins geographic information systems (GIS), supporting applications from hydrology and geomorphology to urban planning and autonomous navigation~\cite{li2004digital}.
The most widely used representation is the digital elevation model (DEM), commonly realized as a regular grid of discrete elevation samples, \ie{}, raster-DEM.
While computationally convenient, raster-DEM imposes inherent limitations: queries between grid points require interpolation; derivative-based quantities such as slope, aspect, and curvature depend on finite-difference operators. Its accuracy is tied to grid resolution and kernel choice~\cite{horn1981hill,zevenbergen1987quantitative}. Additionally, topological analysis on discrete meshes is sensitive to triangulation choices~\cite{de2015morse,feng2024critical}.
Standard raster-DEM workflows usually require separate pipelines for each analysis on the decompressed grid, lacking a unified representation that supports continuous evaluation, well-behaved derivatives, and smooth manifold structure.

Implicit neural representations (INRs) encode a signal as a continuous function $f_\theta\colon (x,y) \mapsto z$ parameterized by a neural network, yielding a compact, resolution-independent, and inherently differentiable model.
ImplicitTerrain~\cite{feng2024implicitterrain} applied this paradigm to DEM data using a cascaded sinusoidal neural network (SIREN) pipeline that separates smooth shape from geometric detail, enabling topographical analysis and Morse-theoretic topological analysis~\cite{milnor1963morse} directly on the learned surface model.
However, ImplicitTerrain's vanilla SIREN architecture has several limitations that prevent it from serving as a practical terrain representation.
First, it lacks \emph{spatial frequency localization}, as SIREN neurons contribute globally, high-frequency components needed only near ridgelines or cliffs also activate in smooth regions, a phenomenon termed \emph{frequency leakage}~\cite{feng2025sasnet}.
Second, its value-only supervision does not exploit the terrain's smooth manifold structure, where well-defined gradients encode physically meaningful shape information.
Moreover, its oversized model ($3 \times 256$ SIREN per stage, $\sim$397K parameters) and multi-scale progressive training further limit its practical deployment. These gaps suggest that terrain INRs are not yet viable as a practical data format.

\begin{figure*}[t]
\centering
\includegraphics[width=\textwidth]{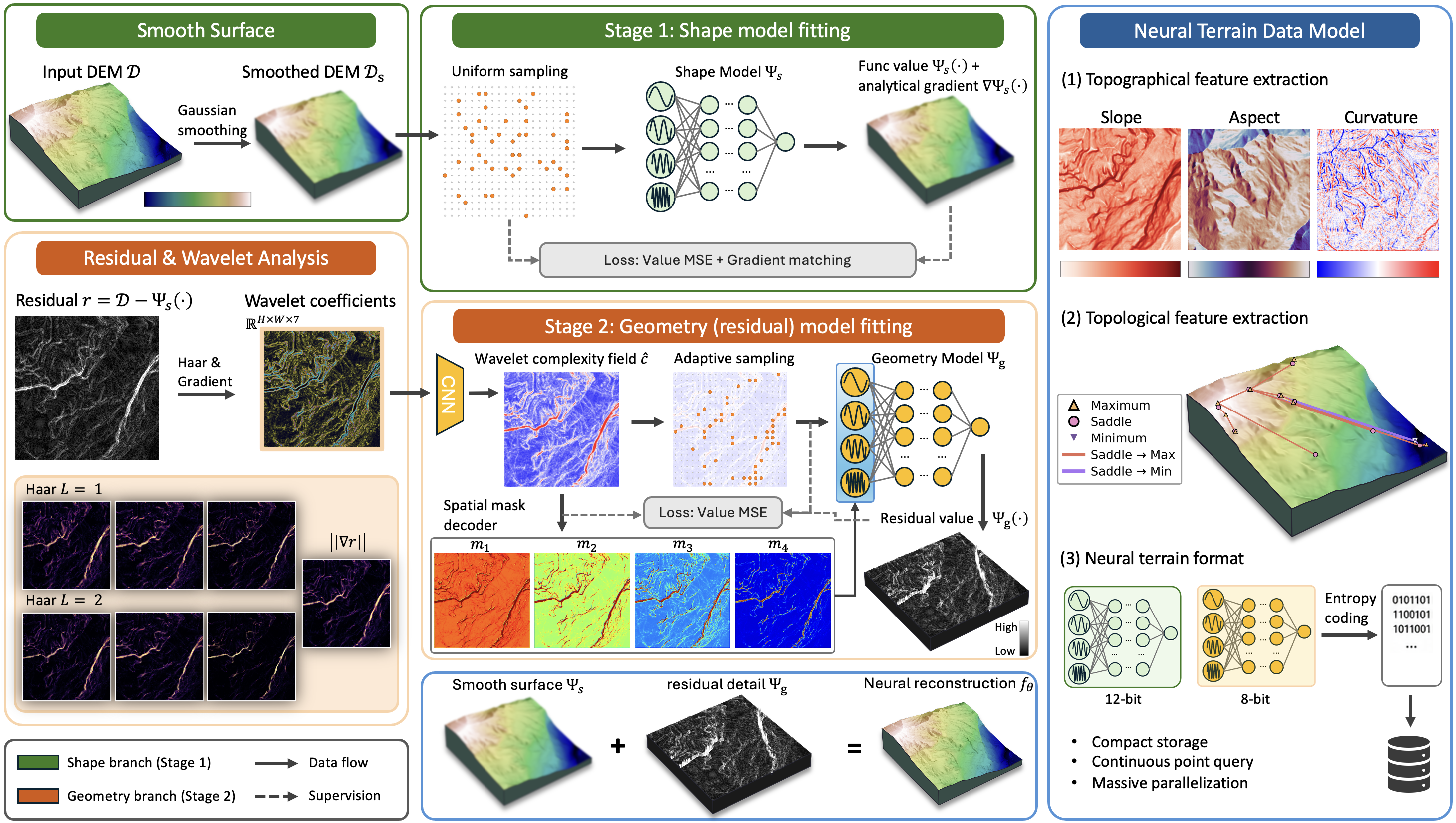}
\caption{Overview of the \method{} pipeline. A terrain DEM is decomposed into a smooth shape model $\psis$ with gradient matching and a residual geometry model $\psig$ with WCF-guided spatial adaptivity. The combined representation supports topological/topographical analysis via analytical gradients, and model compression yields a compact neural terrain data format.}
\label{fig:pipeline}
\end{figure*}

We propose \method{}, shown in Fig.~\ref{fig:pipeline}, which combines wavelet-guided spatial adaptivity, derivative-aware supervision, and post-training model compression to advance terrain INRs toward a compact \emph{neural terrain data format}: a single artifact that couples compact storage, continuous point queries, closed-form derivative evaluation, and resolution-independent reconstruction, targeting offline per-tile encoding settings where these capabilities justify a one-time encoding cost.
ImplicitTerrain's cascaded SIREN architecture is retained, but each per-stage backbone is enhanced through a recent SIREN variant (\ie, TUNER~\cite{novello2024taming}), which provides explicit spectral control. Furthermore, we introduce wavelet-derived spatial masks to localize network frequency components in the spatial domain, as suggested in SASNet~\cite{feng2025sasnet}.
Exploiting the domain gap between RGB images and terrain DEMs, a single scalar complexity field, named \emph{wavelet complexity field} (WCF), is learned from the multi-scale signal analysis results (based on the wavelet analysis).
It exploits the containment hierarchy in the spatial distribution across frequency components, generating ordered masks via learned thresholds and producing an interpretable modulation signal for the SIREN backbone.

A morphologically diverse dataset is curated for the evaluation of \method{}: 50 terrain tiles selected from 33{,}399 swissALTI3D candidates via farthest-point sampling in a morphology and frequency descriptor space, spanning gentle valleys, foothills, and alpine terrain.
Our ablations isolate the improvements from each module proposed in this work, such as network architecture, hyperparameters, and training strategies.
Overall, \method{} achieves 66.25\,dB end-to-end PSNR with 0.13\,m mean absolute error, improving over baseline by +5.70\,dB while using 3.2$\times$ fewer parameters and requiring 55\,s per tile to train on a single GPU.
In rate-distortion comparison, the compressed neural format at 1.23\,bpp is competitive with several established codecs (Quantize+ZSTD~\cite{collet2018zstd}, ZFP~\cite{lindstrom2014zfp}, LERC+ZSTD~\cite{esri2024lerc,collet2018zstd}) in our evaluation.
Although SZ3~\cite{liang2022sz3} remains stronger at matched rates, compared to conventional compressed DEMs, the proposed neural representation additionally provides the following capabilities: a closed-form computation of derivatives for topographical features and topological analysis, massive parallelization on modern GPUs, and resolution-independent reconstruction, without requiring separate processing pipelines.
Our main contributions are:
\begin{itemize}
    \item \textbf{Wavelet complexity field:} an analytically-derived scalar field summarizing local terrain complexity from gradients and multi-scale wavelet coefficients, from which an ordered family of spatial masks is produced via learned thresholds, outperforming learned hash-grid masks on terrain signals.

    \item \textbf{Terrain-aware fitting design:} an adaptation of two established training techniques to the cascaded terrain-INR setting: Hermite-style gradient supervision~\cite{hildebrand1987introduction} that exploits the smooth manifold structure of DEMs to improve both elevation and derivative fidelity, and WCF-guided importance-based sampling that replaces the previous multi-scale progressive training while reducing training cost.

    \item \textbf{Compact neural terrain format:} a system-level contribution that applies mixed-precision quantization and entropy coding to the trained representation and characterizes its accuracy, efficiency, and rate-distortion tradeoff against neural and conventional terrain formats, while retaining INR benefits including continuous surface modeling.
\end{itemize}

\section{Related Work}
\label{sec:related}

\subsection{Implicit Neural Representations}
\label{sec:related:inr}

Implicit neural representations (INRs) parameterize signals as continuous functions $f_\theta\colon \mathbb{R}^d\! \to \! \mathbb{R}^n$ via neural networks, enabling continuous coordinate queries and inherent differentiability~\cite{essakine2024we}.
A central challenge is \emph{spectral bias}: standard ReLU MLPs converge preferentially to low-frequency components, failing to capture fine detail~\cite{Rahaman2018OnTS,Basri2020FrequencyBI,ramasinghe2022frequency}.
SIREN~\cite{sitzmann2020implicit} mitigates this with periodic sinusoidal activations and a dedicated weight initialization, enabling high-fidelity fitting of images, audio, shapes, and their derivatives, while admitting closed-form gradients and Hessians that are analytically tractable~\cite{novello2022exploring}.
Subsequent work broadens spectral coverage via alternative activations such as Gabor wavelets (WIRE)~\cite{saragadam2023wire}, variable-period sinusoids (FINER)~\cite{liu2024finer}, and trainable sinusoidal functions (STAF)~\cite{morsali2025staf}, or use input encoding such as random Fourier features (FFN)~\cite{tancik2020fourier}.
Fourier reparameterization of network weights~\cite{shi2024improved} offers an orthogonal strategy, decomposing weight matrices over fixed Fourier bases without modifying the activation.

A parallel direction augments MLPs with multi-resolution spatial structures.
Instant-NGP~\cite{muller2022instant} couples multi-resolution hash grids with a small MLP for fast training, at the cost of large hash tables and no explicit frequency control.
ACORN~\cite{martel2021acorn} adaptively partitions the spatial domain into individual fitting blocks; BACON~\cite{lindell2022bacon} constrains each layer to a specific frequency band; and MINER~\cite{saragadam2022miner} hierarchically tiles the domain, fitting residuals at progressively finer scales.
Mixture-of-experts variants partition the domain among specialized sub-networks: LoE~\cite{hao2022implicit} tiles position-dependent periodic weights at progressively finer frequencies across MLP layers, while Neural Experts~\cite{ben2024neural} jointly trains a gating network with local experts that dynamically subdivide the domain.

A complementary line targets explicit frequency control and spatial localization.
MFN~\cite{fathony2020multiplicative} composes networks from multiplicative Fourier or Gabor filters, making the representable spectrum directly inspectable.
Y{\"u}ce \etal~\cite{yuce2022structured} formalize a sinusoidal network's representable spectrum as a structured dictionary of sines whose frequencies are integer linear combinations of the first-layer weights; TUNER~\cite{novello2024taming} applies this analysis by freezing SIREN's first layer with weights drawn from a designed spectral distribution, giving explicit Fourier-grounded control.
On the encoding side, spatially-adaptive hash encodings~\cite{walker2025spatially} learn per-region masks over multi-resolution hash-grid features, selecting an effective encoding basis as a function of position.
Beyond spectral control, Mehta \etal~\cite{mehta2021modulated} modulate SIREN's periodic activations with an auxiliary network for generalizable local representations, and INCODE~\cite{kazerouni2024incode} further modulates the SIREN backbone via prior-knowledge embeddings from a task-specific harmonizer.

However, controlling \emph{which} frequencies a network can represent does not determine \emph{where} they are spatially active.
Because SIREN neurons contribute globally, high-frequency components needed only near complex features also activate in smooth regions, a phenomenon termed \emph{frequency leakage}.
SAPE~\cite{hertz2021sape}, built on FFN~\cite{tancik2020fourier}, progressively activates frequency bands per coordinate during training for implicit spatial adaptation, while SASNet~\cite{feng2025sasnet} addresses the SIREN case with spatial masks from a hash-grid MLP that modulate neuron activations via element-wise multiplication to suppress unnecessary high-frequency expression.
While effective on richly textured signals, learned spatial features such as hash grids may be less well matched to low-frequency dominant, low-contrast signals such as terrain residuals, motivating analytically derived complexity measures as an alternative source of spatial adaptivity.

\subsection{Terrain Representation and Analysis}
\label{sec:related:terrain}

The digital elevation model (DEM) is the dominant terrain representation, usually storing elevation as a regular grid~\cite{li2004digital}.
Triangulated irregular networks (TINs)~\cite{garland1995fast} offer adaptive resolution but require explicit mesh construction, while multi-resolution Gaussian pyramids decompose terrain into frequency bands for progressive analysis~\cite{oppenheim1999discrete}.
Common raster- and mesh-based terrain models share a limitation: queries between grid points require interpolation, and derivative quantities depend on the chosen discretization.

Standard topographical features (\eg, slope, aspect, curvature) are computed from DEMs via finite-difference operators~\cite{horn1981hill,zevenbergen1987quantitative}, whose accuracy is inherently tied to grid resolution and kernel size.
Topological analysis provides a complementary, structure-oriented view: Morse theory~\cite{milnor1963morse} extracts critical points, separatrix lines, and the Morse--Smale complex from smooth scalar fields, enabling delineation of ridges, valleys, and drainage divides on terrain surfaces.
For discrete meshes, Forman's discrete Morse theory~\cite{forman1998morse,forman2002user} adapts these concepts to simplicial complexes~\cite{fellegara2014efficient,fellegara2017efficient,fellegara2023terrain}, with persistence diagrams~\cite{fugacci2016persistent} and the Wasserstein distance~\cite{scaramuccia2020computing} providing multi-scale topological summaries.
However, discrete methods are sensitive to mesh construction choices such as diagonal selection in grid triangulation~\cite{de2015morse, feng2024critical}, motivating continuous representations with well-behaved gradients and Hessians to reduce discretization artifacts in topological extraction.

ImplicitTerrain~\cite{feng2024implicitterrain} introduces an INR pipeline for terrain analysis, using a cascaded SIREN in which a surface model fits Gaussian-smoothed data and a geometry model fits the residual, yielding a continuous, differentiable representation that supports Morse-theoretic topological analysis directly on the learned surface.
Concurrently, neural elevation models~\cite{dai2024neural} adapt neural radiance fields to produce a differentiable 2.5D height field from imagery for gradient-based path planning.
However, existing neural terrain representations lack spatial frequency localization, use value-only supervision without exploiting terrain gradient structure, and do not address model compactness for practical storage and distribution.

\subsection{Terrain and Neural Field Data Compression}
\label{sec:related:compression}

DEM data is commonly distributed as GeoTIFF~\cite{mahammad2003geotiff} with integer quantization.
Lossless compression via DEFLATE~\cite{deutsch1996deflate} or Zstandard~\cite{collet2018zstd} yields moderate ratios on floating-point elevation data.
Kidner and Smith~\cite{kidner2003dem} survey DEM-specific techniques including terrain-adapted predictors, and Boucheron and Creusere~\cite{boucheron2005waveletdem} evaluate wavelet-based lossless DEM compression.
In GIS practice, Limited Error Raster Compression~\cite{esri2024lerc} (LERC) provides user-controllable per-pixel error bounds and is widely adopted in operational terrain data pipelines.

Scientific compressors~\cite{cappello2019usecases} optimize different objectives.
Transform coding-based methods form one family: ZFP~\cite{lindstrom2014zfp} uses orthogonal block transforms with embedded coding for fixed-rate or fixed-accuracy control, while SPERR~\cite{li2023sperr} pairs CDF 9/7 wavelets with SPECK coding.
Prediction-based methods form another: SZ~\cite{tao2019szzfp,liang2022sz3} excels on spatially smooth data, and FPZIP~\cite{lindstrom2006fpzip} extends Lorenzo prediction to lossless scientific arrays.
HPEZ~\cite{liu2024hpez} advances this line with auto-tuned multi-component interpolation.
Closer to terrain, Xie \etal~\cite{xie2010slope} propose \emph{slope-preserving} lossy compression that prioritizes gradient accuracy over absolute elevation.
Although these formats enable efficient storage and, in some cases, block-level random access, their outputs are values alone: gradient computation or topological extraction is left to separate processing pipelines.

Encoding raw data samples as neural network weights and compressing those weights is an emerging paradigm.
COIN~\cite{Dupont2021COINCW} and COIN++~\cite{dupont2022coin++} demonstrate modulation-based INR compression with meta-learning, while NIRVANA~\cite{maiya2023nirvana} applies quantization-aware training to video INRs.
Cool-Chic~\cite{ladune2023coolchic} achieves competitive image compression with lightweight neural decoders, and BRIEF~\cite{chen2023brief} applies INRs for biomedical data.
Standard weight compression techniques, including post-training quantization (PTQ), network pruning, and entropy coding~\cite{dupont2022coin++,Dupont2021COINCW}, provide complementary rate-distortion trade-offs.
Unlike traditional formats, a compressed neural representation can retain functional capabilities such as continuous queries and analytical derivative computation, though this potential has not been systematically explored for terrain data.

In summary, prior work separately addresses spectral control and spatial adaptivity in INRs, derivative-based supervision in neural fields, continuous terrain modeling, and compression of rasters or neural network weights.
In this work, we bring these threads together in a single terrain representation. Besides, we propose a localized frequency control through a wavelet-derived complexity field, incorporate derivative-aware training tailored to terrain, and apply compact quantized storage, while supporting continuous evaluation and derivative queries directly from the stored representation.

\section{Method}
\label{sec:method}

We assume the underlying terrain surface is a continuous function $f: \mathbb{R}^2 \to \mathbb{R}$ that maps normalized coordinates $(x,y) \in [0,1]^2$ to elevation $z$, of which a terrain tile DEM $\mathcal{D} \in \mathbb{R}^{H \times W}$ is a discrete sampling at grid points. We seek a neural approximation $f_\theta$ of $f$ whose grid-point values reconstruct $\mathcal{D}$ and that provides resolution-independent continuous point queries, analytical differentiability, and continuous shape analysis without discretization.
Figure~\ref{fig:pipeline} illustrates the full \method{} pipeline.

\subsection{Network Architecture}
\label{sec:method:arch}

Our terrain representation is a cascade of two frequency-controlled SIREN networks with input-layer spatial masking.
A \emph{shape model} $\psis$ fits a Gaussian-smoothed version of the DEM $D_s$ and captures the large-scale smooth manifold, while a \emph{geometry model} $\psig$ fits the full-resolution residual $r = \mathcal{D} - \psis(\cdot)$ and recovers fine-scale detail; the final reconstruction is $f_{\theta}(x,y) = \psis(x,y) + \psig(x,y)$.

Each stage is a SIREN~\cite{sitzmann2020implicit} tailored to terrain fitting by two design choices: (i) for $\psis$ and $\psig$, a \emph{frozen first layer} that acts as a designed Fourier basis, providing explicit spectral control over the network's representable frequencies, and (ii) for $\psig$, an input-layer \emph{spatial mask} derived from a wavelet-analytic complexity signal, localizing those frequencies to the terrain regions that need them.
The SIREN backbone is retained because any derivative of a sinusoid remains a sinusoid, so the network yields smooth and analytically tractable derivatives at all orders, a property important for the topographical and topological terrain analysis we target~\cite{novello2022exploring,feng2024implicitterrain}.

Let $\mathbf{x} = (x,y) \in [0,1]^2$ be the normalized input coordinate and $L$ the number of hidden layers.
Each stage computes:
\begin{equation}
\begin{aligned}
    \mathbf{h}_1(\mathbf{x}) &= \sin\bigl(\omega_0\,(\mathbf{W}_0\,\mathbf{x} + \mathbf{b}_0)\bigr) \odot \mathcal{M}(\mathbf{x}), \\
    \mathbf{h}_{l+1}(\mathbf{x}) &= \sin\bigl(\omega_0\,(\mathbf{W}_l\,\mathbf{h}_l + \mathbf{b}_l)\bigr), \quad l = 1,\ldots,L-1, \\
    f_\theta(\mathbf{x}) &= \mathbf{W}_L \mathbf{h}_L + b_L,
\end{aligned}
\label{eq:siren}
\end{equation}
where $\omega_0$ is the frequency scaling factor, $\{\mathbf{W}_l, \mathbf{b}_l\}_{l=0}^{L}$ are the layer weights and biases, and $\odot$ denotes element-wise multiplication between the first-layer activations and a per-neuron spatial mask $\mathcal{M}(\mathbf{x}) \in [0,1]^n$, with $n$ the width of the first layer.
The same formulation covers both stages: the geometry model uses a non-trivial WCF-derived mask (Sec.~\ref{sec:method:wcf}), whereas the shape model sets $\mathcal{M}(\mathbf{x}) \equiv \mathbf{1}$, since its smoothed target contains no spatially localized high-frequency content to gate, so Eq.~\ref{eq:siren} reduces to a standard frequency-embedded SIREN for $\psis$.
The first-layer parameters $\mathbf{W}_0 \in \mathbb{R}^{n \times 2}$ and $\mathbf{b}_0 \in \mathbb{R}^n$ are frozen at initialization and determine the representable spectrum of the network, while $\mathcal{M}$ localizes that spectrum in space.
All other weights and biases are trained jointly.

Y{\"u}ce \etal~\cite{yuce2022structured} showed that every hidden neuron of a SIREN can be expanded as a sum of sines whose frequencies are integer linear combinations of the rows of $\mathbf{W}_0$: for a 2D input, the set $\{\mathbf{k}^\top \mathbf{W}_0 : \mathbf{k} \in \mathbb{Z}^n\}$ defines the network's representable 2D spectrum.
This recasts the first layer as a frequency embedding whose frequencies, motivated by the Fourier analysis of bandlimited 2D elevation signals, can be sampled directly on the integer frequency lattice.
Following TUNER~\cite{novello2024taming} and SASNet~\cite{feng2025sasnet}, we set the rows of $\mathbf{W}_0$ to 2D frequency vectors $\boldsymbol{\omega}_i \in \mathbb{Z}^2$ drawn from an integer frequency grid, and we sample the phases $\mathbf{b}_0$ uniformly in $[0, 2\pi)$.
The $n$ rows are partitioned into $K+1$ frequency bands by magnitude: band $0$ collects the low-frequency rows satisfying $\|\boldsymbol{\omega}_i\|_\infty \leq \mathcal{L}$, and bands $1,\ldots,K$ cover the higher-frequency region from the low-range $\mathcal{L}$ up to a bandlimit $\mathcal{B}$, with each band corresponding to a contiguous range of frequencies.
This construction makes the representable spectrum explicit and reproducible, rather than leaving it to emerge implicitly from random initialization as in standard SIREN.

For spatially-adaptive masking, let $\mathbf{W}_0^{(i)}$ denote the rows of $\mathbf{W}_0$ assigned to frequency band $i$, with $i = 0$ the low-frequency band and $i = 1,\ldots,K$ the higher-frequency bands.
Given per-band scalar masks $m_i(\mathbf{x}) \in [0,1]$, whose construction from the wavelet complexity field is described in Sec.~\ref{sec:method:wcf}, we form the per-neuron mask $\mathcal{M}$ by tiling $m_i(\mathbf{x})$ across every neuron in band $i$:
\begin{equation}
    \mathcal{M}(\mathbf{x}) = \bigl[\,\underbrace{m_0(\mathbf{x}),\ldots,m_0(\mathbf{x})}_{|\mathbf{W}_0^{(0)}|},\; \underbrace{m_1(\mathbf{x}),\ldots,m_1(\mathbf{x})}_{|\mathbf{W}_0^{(1)}|},\; \ldots\,\bigr]^{\!\top},
    \label{eq:mask_tiling}
\end{equation}
and apply it element-wise to the first-layer activations as in Eq.~\ref{eq:siren}.
The low-frequency band is always active ($m_0 \equiv 1$); only the $K$ higher-frequency bands are gated. In practice, spatial masks are applied to a group of neurons through broadcasting in PyTorch for efficiency.
This band-wise masking structure is adopted from SASNet~\cite{feng2025sasnet}; our departure lies in how $\{m_i\}$ are produced.
We replace SASNet's jointly trained hash-grid branch, which tends to collapse on terrain residuals (Sec.~\ref{sec:method:wcf}), with a \emph{wavelet complexity field} that derives $\{m_i\}$ from analytically computed wavelet coefficients of the input terrain, providing an interpretable mask signal grounded in the multi-scale frequency content of the data itself.
Masks are applied only at the first layer; hidden-layer masks, used by SASNet for images, degrade terrain fitting and are therefore omitted (see Supp.~\ref{sec:supp:hidden_masks}).

\subsection{Wavelet Complexity Field (WCF)}
\label{sec:method:wcf}

WCF produces the geometry model's $K$ per-band spatial masks $\{m_i\}$ from a single scalar complexity field, computed by a lightweight CNN decoder over analytical multi-scale wavelet features of the input and sliced into $K$ nested masks by $K$ strictly ordered thresholds.
The masks gate the geometry model's first-layer frequencies (Eq.~\ref{eq:siren}), localizing high-frequency capacity to complex terrain regions.

\begin{figure}[t]
\centering
\includegraphics[width=\columnwidth]{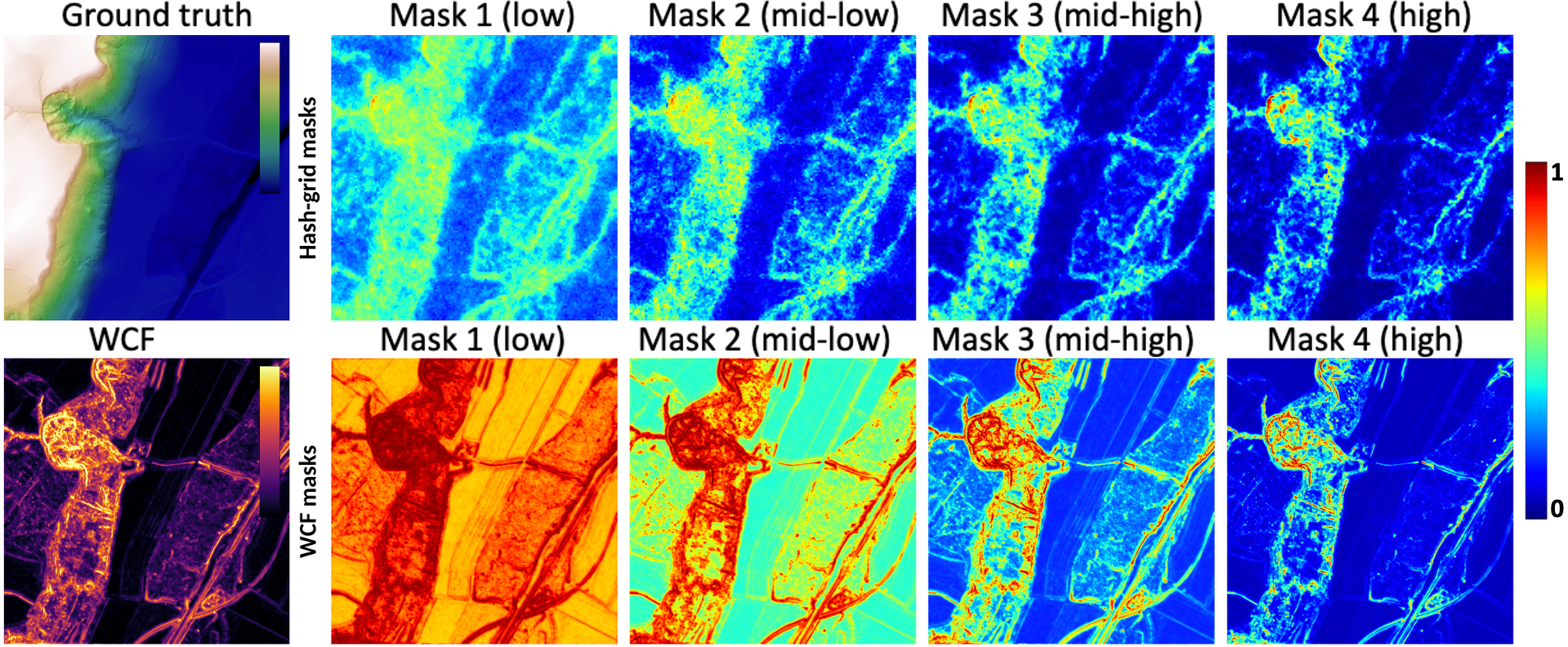}
\caption{Mask comparison on a terrain tile. Hash-grid masks (top) are spatially noisy and lack interpretability, failing to distinguish complex from simple regions. WCF masks (bottom) show a clear spatial structure with a containment hierarchy from low to high frequency bands.}
\label{fig:hashgrid_failure}
\vspace{-0.4em}
\end{figure}

Wavelet coefficients decompose a 2D signal into multi-scale frequency components while preserving spatial localization, with large-magnitude coefficients indicating rapid spatial variation at that scale and location~\cite{mallat1989multiresolution,boucheron2005waveletdem}.
Computing them analytically from the input terrain makes the complexity signal available before optimization. SASNet's hash-grid masks, by contrast, have no analytical form and are jointly trained with the SIREN under the geometry model's residual reconstruction loss; since this residual is itself the geometry model's fitting target, the masks must be inferred from the same smooth, low-contrast signal they are meant to gate, which causes them to degrade on post-shape residuals (Fig.~\ref{fig:hashgrid_failure}).

To turn the complexity field into the $K$ per-band masks, we exploit the \emph{containment hierarchy} reported by SASNet~\cite{feng2025sasnet} for well-learned image masks: the active region of each higher-frequency band is contained in that of the lower-frequency bands, since regions requiring fine detail often also require coarser content. We find this hierarchy is also important for stable terrain optimization.
We therefore derive all $K$ masks from the WCF (a single scalar field) through $K$ ordered thresholds, which ensures containment by construction and reduces the decoder to a scalar-valued regression.

We compute the stationary wavelet transform (SWT)~\cite{nason1995stationary} of residual $r$ at $L = 2$ levels with Haar wavelets; the SWT is shift-invariant and preserves spatial resolution at every level by omitting downsampling.
This yields six absolute-valued detail coefficients $\{|d^{(\ell)}_h|, |d^{(\ell)}_v|, |d^{(\ell)}_d|\}_{\ell=1}^{L}$ encoding horizontal, vertical, and diagonal frequency content at each coordinate.
We append the local gradient magnitude $\|\nabla r\|$ as a seventh first-order channel, stack into $\mathbf{F}(r) \in \mathbb{R}^{7 \times H \times W}$, and z-score normalize each channel to obtain $\bar{\mathbf{F}}(r)$.

A CNN decoder $g_\phi$ then maps $\bar{\mathbf{F}}(r)$ to a scalar field on a low-resolution grid, which is per-tile instance normalized to obtain the complexity field $\hat{c}$:
\begin{equation}
    c = g_\phi\bigl(\bar{\mathbf{F}}(r)\bigr), \qquad \hat{c}(\mathbf{x}) = {(c(\mathbf{x}) - \mu_c)}/{(\varsigma_c + \epsilon)},
    \label{eq:wcf}
\end{equation}
where $\mu_c, \varsigma_c$ are the spatial mean and standard deviation of $c$ over the tile.
The two normalizations play complementary roles: the per-channel z-scoring of $\mathbf{F}(r)$ balances the magnitudes of the wavelet feature channels before the decoder, while the per-tile instance normalization of $c$ standardizes the threshold operating range so a single learned set of $\{\tau_i\}$ generalizes across tiles with different elevation ranges.

\begin{figure}[t]
\centering
\includegraphics[width=\columnwidth]{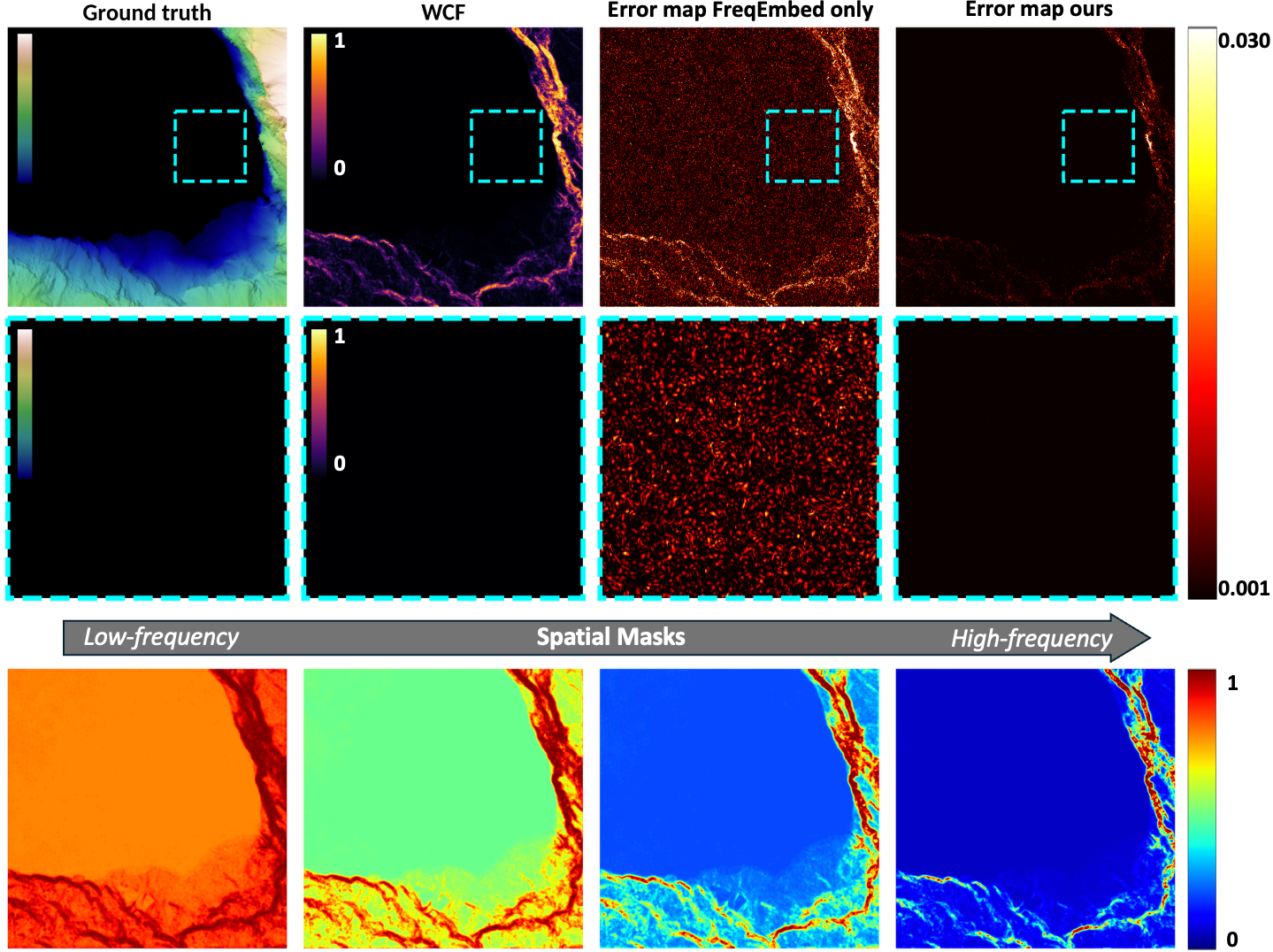}
\caption{Frequency leakage in the geometry model. Without spatial masks (bottom row), high-frequency neurons introduce spurious oscillations in flat regions. WCF masks (second column) localize errors to complex terrain. The middle row zooms into a flat region (cyan box), where WCF masks reduce mean error by 80.5\%.}
\label{fig:freq_leakage}
\vspace{-0.4em}
\end{figure}

We map $\hat{c}$ to $K$ soft per-band masks through $K$ strictly ordered thresholds $\tau_1 < \cdots < \tau_K$:
\begin{equation}
    m_i(\mathbf{x}) = \sigma\bigl(\hat{c}(\mathbf{x}) - \tau_i\bigr),
    \label{eq:masks}
\end{equation}
where $\sigma$ is the sigmoid. To enforce strict ordering without constrained optimization, we adopt a cumulative-softplus reparameterization:
\begin{equation}
    \tau_1 = \tilde{\tau}_1, \qquad \tau_i = \tau_{i-1} + \mathrm{softplus}(\delta_i), \quad i = 2, \ldots, K,
    \label{eq:threshold_reparam}
\end{equation}
where $\mathrm{softplus}(\delta_i) = \log(1 + e^{\delta_i})$, and $\{\tilde{\tau}_1, \delta_2, \ldots, \delta_K\}$ are learnable scalars that parameterize the thresholds $\{\tau_i\}$ filtering $\hat{c}(\mathbf{x})$: $\tilde{\tau}_1$ sets the lowest threshold, and each $\delta_i$ controls the gap between consecutive thresholds. Each step adds a positive $\mathrm{softplus}(\delta_i)$, so $\tau_i > \tau_{i-1}$ by construction.
Higher thresholds suppress more locations, so the active regions of the masks form a nested chain: the lowest band is active almost everywhere, the highest only on the most complex features.
The $K$ masks are tiled across the corresponding neuron groups of $\mathbf{W}_0$ via Eq.~\ref{eq:mask_tiling}.

Spatial masking is essential because SIREN neurons act globally over the input domain: without it, high-frequency components needed only in complex regions produce spurious oscillations in flat areas (Fig.~\ref{fig:freq_leakage}).
Empirically (Sec.~\ref{sec:exp:fitting}), the frequency embedding alone yields no improvement on the geometry stage, while adding WCF masks recovers $+2.64$\,dB and reduces mean error in the highlighted flat region by 80.5\%.
WCF masks also display the expected containment hierarchy (Fig.~\ref{fig:hashgrid_failure}), with mean activation decreasing monotonically from $0.82$ in the lowest band to $0.13$ in the highest, concentrating high-frequency capacity on ridgelines and sharp slope transitions.

\subsection{Gradient Matching}
\label{sec:method:gradient}

Supervising both function values and their derivatives, known as Sobolev training~\cite{czarnecki2017sobolev}, improves data efficiency and generalization in neural networks, and has been applied in INRs through Eikonal constraints for signed distance functions~\cite{gropp2020implicit} and approximated image derivatives~\cite{yuan2022sobolev}.
However, derivative supervision remains uncommon in general neural field training, as many signals lack well-defined or easily computable ground-truth gradients.
Terrain DEMs are a notable exception: the elevation function over a smooth manifold has first-order partial derivatives $(\partial f_{\theta}/\partial x, \partial f_{\theta}/\partial y)$ that correspond directly to physical slope components, are computable from the DEM via finite differences, and are essential for downstream geospatial analysis.
We exploit this structure through \emph{gradient matching}, a Hermite-style~\cite{hildebrand1987introduction} loss $\mathcal{L}_\text{MSE} + \lambda \cdot \mathcal{L}_\text{grad}$ that supervises both function values and derivatives:
\begin{align}
    \mathcal{L}_\text{MSE} &= \frac{1}{N}\sum_{i=1}^N \bigl(f_\theta(x_i, y_i) - f_i\bigr)^2 \\
    \mathcal{L}_\text{grad} &= \frac{1}{M}\sum_{j=1}^M \left\| \nabla f_\theta(x_j, y_j) - \nabla f_j \right\|^2
    \label{eq:grad_loss}
\end{align}
The gradient targets $\nabla f_j$ are pre-computed from the DEM using central finite differences, and $\nabla f_\theta$ is obtained via automatic differentiation through the SIREN network.
We set $\lambda = 0.1$ to keep the gradient term on a comparable scale to $\mathcal{L}_\text{MSE}$ on normalized elevation, and randomly subsample $M = 10{,}000$ coordinates per batch for gradient supervision to limit computational overhead (see Sec.~\ref{sec:supp:gm_cost} for cost analysis).

Gradient matching increases the effective supervision density: each training coordinate provides one value constraint and two first-order derivative constraints, adding complementary geometric information per sample.
For terrain, these derivative constraints are particularly informative because the elevation surface is smooth and its gradients encode physically meaningful local geometry (\ie, slope direction and magnitude) that pure MSE fitting loss must infer indirectly from neighboring point values.

Gradient matching also complements the frequency embedding layer: while the frequency embedding broadens the network's \emph{spectral} capacity by covering more Fourier frequencies, gradient supervision provides denser \emph{spatial} constraints that guide the network to allocate these components accurately across the terrain surface.
We apply gradient matching only to the shape model $\psis$, because it fits the smooth manifold where gradients are spatially coherent and physically meaningful (\eg, terrain slope).
Together, as shown in Sec.~\ref{sec:exp:ablation}, the two mechanisms address frequency coverage and spatial fidelity from complementary directions.

\subsection{Efficient Training and Inference}
\label{sec:method:efficiency}

A practical neural terrain format must be efficient to train and to query.
We address training cost from three aspects: (i) adaptive coordinate sampling, (ii) analytical gradient inference that leverages SIREN's differentiable structure as a built-in capability of the representation, and (iii) mixed-precision computation with kernel compilation that accelerates both training and inference.

\vspace{0.2em}\noindent\textbf{Adaptive sampling.}
Previous work (ImplicitTerrain) training uses all grid coordinates, treating all spatial locations equally at each iteration.
For terrain, this is inefficient because in flat regions, the model converges faster and does not benefit from dense sampling, while in complex regions (\eg, ridgelines, cliffs), it requires concentrated optimization effort.

We use different sampling strategies for the two stages, matched to the spectral character of each target.
For the shape model, whose target is the band-limited smoothed manifold, we apply simple uniform random subsampling of the $500 \times 500$ grid at each iteration.
For the geometry model, whose target is the high-frequency residual with strongly non-uniform spatial importance, we exploit the normalized WCF complexity field $\hat{c}(x,y)$ as an importance map and sample training coordinates from a mixture distribution:
\begin{equation}
    p(x,y) = (1-\alpha) \cdot \mathcal{U} + \alpha \cdot \hat{c}^{+}(x,y) / {\sum} \hat{c}^{+},
    \label{eq:adaptive_sampling}
\end{equation}
where $\hat{c}^{+} = \max(\hat{c}, \epsilon_s)$ is the rectified normalized complexity (with $\epsilon_s > 0$ ensuring minimum coverage of all regions), $\mathcal{U}$ is the uniform distribution over the grid, and $\alpha = 0.75$ controls the adaptive fraction.
Sample indices are drawn per iteration on the CPU in parallel with GPU training to avoid sampling overhead.

Empirically, both strategies provide substantial speedups at negligible quality cost.
For the shape model, $25\%$ uniform subsampling incurs only $-0.01$\,dB PSNR loss while providing a $1.84\times$ training speedup (Supp.~\ref{sec:supp:shape_as}); the smooth manifold tolerates aggressive subsampling because the target is band-limited and spatially redundant.
For the geometry model, complexity-guided sampling slightly improves fitting quality ($+0.53$\,dB) while being $2.25\times$ faster, as the network concentrates optimization effort on regions that need it most.
Combined with the cascaded architecture, adaptive sampling also removes the need for ImplicitTerrain's multi-scale progressive training: a single-resolution pass suffices and yields $+3.42$\,dB higher end-to-end PSNR at $1.57\times$ faster wall-clock time (Supp.~\ref{sec:supp:multiscale}).

\vspace{0.2em}\noindent\textbf{Analytical gradient queries.}
At inference time, WCF masks are precomputed and frozen, so the model reduces to a masked SIREN with fixed per-location gains.
A key advantage of this model is that spatial derivatives can then be computed analytically via the chain rule through SIREN layers, without constructing a computational graph~\cite{novello2022exploring}.
For a SIREN hidden layer $\mathbf{h}_{l+1} = \sin(\omega_0 (\mathbf{W}_l \mathbf{h}_l + \mathbf{b}_l))$, the Jacobian with respect to the layer input is:
\begin{equation}
    \frac{\partial \mathbf{h}_{l+1}}{\partial \mathbf{h}_l} = \omega_0 \cdot \mathrm{diag}\bigl(\cos(\omega_0 (\mathbf{W}_l \mathbf{h}_l + \mathbf{b}_l))\bigr) \cdot \mathbf{W}_l.
    \label{eq:analytical_grad}
\end{equation}
To avoid the overhead of PyTorch's autograd backward, we implement Eq.~\ref{eq:analytical_grad} explicitly. The forward pass caches each layer's cosine pre-activation $\cos(\omega_0(\mathbf{W}_l \mathbf{h}_l + \mathbf{b}_l))$. A manual chain-rule pass propagates the running 2D input gradient through the frozen weights $\{\mathbf{W}_l\}$, scaled element-wise by the cached cosines, producing $\partial f_\theta/\partial x$ and $\partial f_\theta/\partial y$ alongside $f_\theta$ in a single call.
At equal FLOP count to autograd, this manual chain has no graph-construction or dispatch overhead and can be fused end-to-end into a single GPU kernel.
These analytical derivatives can be evaluated at any continuous coordinate without resampling or grid construction, and directly support the downstream topographical and topological analysis in Sec.~\ref{sec:exp:analysis}.

\vspace{0.2em}\noindent\textbf{Mixed-precision computation and compilation.}
We accelerate both training and inference using automatic mixed precision (AMP)~\cite{micikevicius2018mixed} and kernel compilation via \texttt{torch.compile}~\cite{ansel2024pytorch}.
AMP performs most arithmetic in float16 while keeping master weights and numerically sensitive operations (\eg, reductions, normalization) in float32, reducing memory bandwidth and compute cost without sacrificing training stability.
Kernel compilation fuses operation sequences into optimized GPU kernels, substantially reducing Python interpreter overhead and the cost of managing the autograd computational graph.
This compilation step is important for analytical gradient queries: the analytical formulation has the same FLOP count as autograd, and without compilation the speedup is marginal ($1.06\times$); with compilation, the fused forward-and-gradient kernel achieves $4.63\times$ speedup, reaching 77.3\,M queries/s for simultaneous elevation and gradient evaluation (Sec.~\ref{sec:exp:compression}).

\subsection{Model Compression}
\label{sec:method:compression}

For the neural terrain representation to be practical as a data format, the trained weights must be compressed for storage and distribution.
We apply a standard two-stage neural-weight compression pipeline~\cite{gholami2022survey}: post-training quantization (PTQ) of the float32 weights to low-bit integers, followed by lossless entropy coding of the quantized symbol stream.
Both stages operate on the trained model and add no per-tile training cost.

PTQ converts the trained float32 weights to a lower-precision integer representation after training, without any additional retraining loop~\cite{gholami2022survey}.
For a weight tensor $\mathbf{W}$ at bit-width $b$, we compute a scale $s = \max(|\mathbf{W}|)/(2^{b-1}-1)$ and round each entry to the nearest signed $b$-bit integer $\mathbf{W}_{\mathrm{int}}$; at inference the integer weights are dequantized as $\hat{\mathbf{W}} = s\,\mathbf{W}_{\mathrm{int}}$ for computation, while only $\mathbf{W}_{\mathrm{int}}$ and $s$ are stored on disk.
We use symmetric uniform weight-only quantization, with per-output-channel scales for weights and per-tensor scales for biases; the frozen frequency-embedding layer $\mathbf{W}_0$ is reproducible from its initialization seed and is not stored.
A per-layer sensitivity sweep (Supp.~\ref{sec:supp:quant_sensitivity}) shows that the two cascade stages have different bit-width tolerances: the shape model $\psis$ requires $b=12$ to preserve gradient accuracy for downstream topographical and topological analysis, whereas the geometry model $\psig$ and the WCF decoder tolerate $b=8$ with negligible PSNR loss.
We therefore adopt a mixed-precision configuration of $b=12$ for $\psis$ and $b=8$ for $\psig$ and the WCF decoder.

After PTQ, the integer weight values are not uniformly distributed, with low-magnitude values far more common than extreme ones, leaving headroom for additional lossless compression.
We compress the quantized weight stream with arithmetic coding~\cite{witten1987arithmetic}, which encodes a sequence of symbols as a single fractional number whose code length approaches the empirical entropy of the symbol distribution.
Beyond the network weights, the WCF mask generator at inference also requires the per-tile complexity field $\hat{c}$ that produces the band masks through the learned thresholds.
We store this field at the low resolution at which it is produced (the same grid used for bilinear interpolation during training), uniformly quantized to 4 bits and entropy-coded with the same arithmetic coder.
The stored artifact per tile thus consists of the entropy-coded integer weight stream, per-channel scales, the learned thresholds and normalization statistics, and the entropy-coded low-resolution complexity field.

We use PTQ over quantization-aware training (QAT), as our ablation (Supp.~\ref{sec:supp:qat}) shows that QAT yields negligible quality improvement while substantially complicating training.
Rate-distortion results, including comparisons against established codecs, are reported in Sec.~\ref{sec:exp:compression}.

\section{Experiments}
\label{sec:experiments}

\begin{table*}[t]
\centering
\small
\caption{Main fitting results on 50 tiles, best per column in \textbf{bold}. E2E PSNR (mean $\pm$ std) is computed on the full cascaded reconstruction. MAE and MaxAE are scaled back to the original elevation scale. Peak Mem is the peak GPU memory during training, and Geo-Iter is the per-iteration wall-clock time of the geometry model, which is the training bottleneck.}
\label{tab:main_results}
\vspace{-0.8em}
\begin{tabular}{lcccccc}
\toprule
Method & Params & E2E PSNR (dB) $\uparrow$ & MAE (m) $\downarrow$ & MaxAE (m) $\downarrow$ & Peak Mem (GB) $\downarrow$ & Geo-Iter (ms) $\downarrow$ \\
\midrule
ImplicitTerrain~\cite{feng2024implicitterrain}  & 397K & 60.55 $\pm$ 5.67 & 0.197 & 29.73 & 9.8 & 81.9 \\
ImplicitTerrain-128                              & 100K & 49.74 $\pm$ 5.34 & 1.436 & 32.46 & 4.9 & 34.5 \\
SASNet~\cite{feng2025sasnet}            & 139K & 64.48 $\pm$ 4.27 & \textbf{0.125} & 17.56 & 5.4 & 41.3 \\
\textbf{\method{}}                               & 124K & \textbf{66.25 $\pm$ 4.09} & 0.132 & \textbf{11.85} & \textbf{1.7} & \textbf{20.2} \\
\bottomrule
\end{tabular}
\end{table*}

\subsection{Experimental Setup}
\label{sec:exp:setup}

For evaluation, we curated a 50-tile terrain dataset from SwissTopo's swissALTI3D~\cite{swissALTI3D}, a high-resolution LiDAR-derived DEM of Switzerland, selecting tiles from 33{,}399 candidates via farthest-point sampling in a 10-dimensional morphology and frequency descriptor space to maximize morphological diversity.
Each processed tile covers $1000 \times 1000$ pixels at 1\,m resolution.
The resulting set spans alpine ridges, glacial valleys, plateaus, and mixed terrain, with elevation ranges from 101\,m to 1{,}668\,m.
Details of the curation pipeline are provided in Supp.~\ref{sec:supp:dataset}.

We compare \method{} with three neural terrain representations:
(1)~\textbf{ImplicitTerrain}~\cite{feng2024implicitterrain}, a cascaded SIREN baseline with three 256-unit hidden layers and $\omega_0 = 30$;
(2)~\textbf{ImplicitTerrain-128}, the same baseline reduced to a $3 \times 128$ SIREN backbone for an architecture-matched comparison; and
(3)~\textbf{SASNet}~\cite{feng2025sasnet}, which augments a $3 \times 128$ SIREN backbone with multi-scale hash-grid masks.
For compression evaluation, we additionally compare against several widely-used classical terrain formats (see Sec.~\ref{sec:exp:compression}).

Both stages of \method{} use $L = 3$ hidden SIREN layers with width $128$.
The shape model uses $\omega_0 = 30$ with a low-frequency-only frequency embedding ($\mathcal{L} = 10$, $K = 0$), matching the spectral content of the smoothed terrain; no spatial masking is applied ($\mathcal{M} \equiv \mathbf{1}$).
The geometry model uses $\omega_0 = 150$ with a 50\%/50\% low/high-frequency neuron split ($\mathcal{L} = 6$, $K = 4$ higher-frequency bands up to bandlimit $\mathcal{B} = 40$); its WCF masks gate the $K$ higher-frequency bands per spatial location and are produced from Haar SWT at two decomposition levels, decoded by a 3-layer CNN ($7 \to 48 \to 48 \to 1$).
This configuration reduces ImplicitTerrain's $3 \times 256$ backbone, yielding a more compact neural terrain format.
Model sizes were selected through a preliminary search to balance parameter count, training/inference speed, and fitting quality.

All models are optimized with Adam~\cite{kingma2014adam} at a learning rate of $10^{-4}$.
The shape model trains for 3{,}000 iterations on $500 \times 500$ Gaussian-smoothed terrain ($\sigma = 4.0$), and the geometry model trains for 2{,}000 iterations on $1000 \times 1000$ normalized residuals; the per-stage $\omega_0$ values above were selected through hyperparameter search.
Gradient matching ($\lambda = 0.1$, $M = 10{,}000$ subsampled coordinates) and adaptive sampling ($\alpha = 0.75$, geometry model only) are applied to \method{} and, where compatible, to SASNet; the SIREN-based ImplicitTerrain baselines retain their original training recipe without these additions.
Training runs with \texttt{torch.compile} and AMP on a single NVIDIA RTX A5000 GPU.

We report peak signal-to-noise ratio (PSNR) computed on per-tile min-max-normalized elevations in $[0,1]$, mean absolute error (MAE) and maximum absolute error (MaxAE) in meters after rescaling to each tile's original elevation range, and per-iteration wall-clock time to characterize training throughput.
For the shape model, we additionally report gradient MAE (GradMAE), defined as
\begin{equation}
\mathrm{GradMAE} = \frac{1}{N}\sum_{i=1}^{N} \bigl\| \nabla \psis(\mathbf{x}_i) - \widehat{\nabla f}(\mathbf{x}_i) \bigr\|_1,
\end{equation}
where $\widehat{\nabla f}$ denotes the ground-truth gradient obtained via central finite differences on the DEM and the average is taken over all $N$ grid coordinates.
GradMAE quantifies derivative fidelity, which is essential for the downstream terrain analysis tasks in Sec.~\ref{sec:exp:analysis}.

\subsection{Terrain Fitting Results}
\label{sec:exp:fitting}

Table~\ref{tab:main_results} compares \method{} against baseline methods on all 50 tiles.
We report end-to-end (E2E) PSNR, computed on the full cascaded reconstruction $\hat{z} = \psis + \psig$ against the ground-truth DEM, so it reflects the combined fidelity of the shape and geometry stages rather than either stage in isolation.
\method{} achieves 66.25\,dB E2E PSNR, improving over ImplicitTerrain by +5.70\,dB at $3.2\times$ fewer parameters and $5.8\times$ lower peak GPU memory.
Against SASNet, \method{} gains +1.77\,dB and reduces worst-case error by 33\% (11.85 vs.\ 17.56\,m) while running $2.0\times$ faster per iteration.
SASNet attains a slightly lower MAE (0.125 vs.\ 0.132\,m), likely a consequence of its larger auxiliary masking module (39K vs.\ 24K parameters for WCF) that can fit localized outliers more aggressively; we prioritize PSNR and MaxAE as the primary metrics since they better reflect worst-case fidelity for downstream analysis.

\begin{figure}[t]
\centering
\includegraphics[width=\columnwidth]{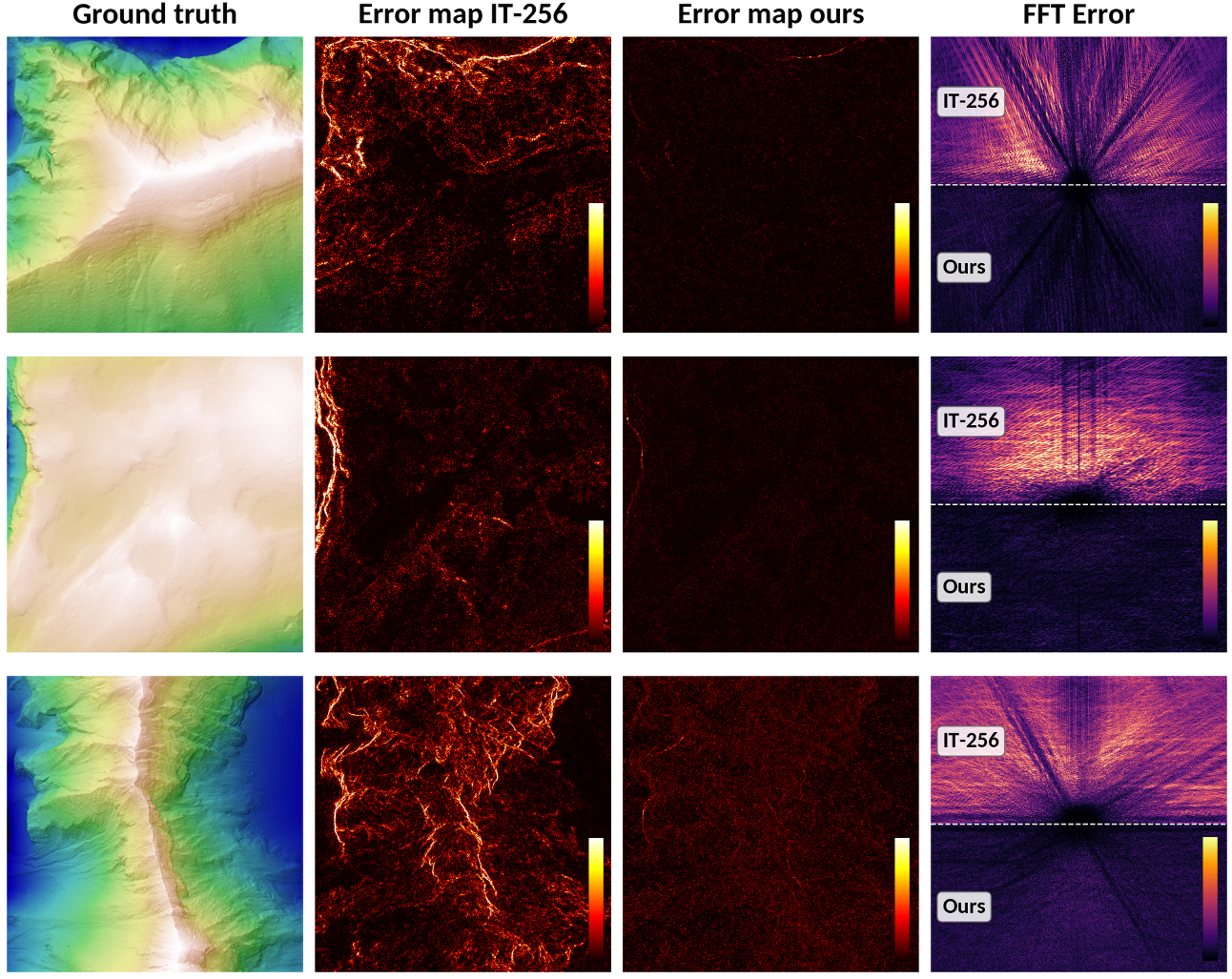}
\caption{Qualitative comparison on three terrain tiles. The 2nd and 3rd columns show spatial error maps for ImplicitTerrain and \method{} (hot colormap for the absolute error up to 1.25\% of the input elevation), while the 4th column shows the frequency-domain error map comparison.}
\label{fig:qualitative}
\vspace{-0.4em}
\end{figure}

Figure~\ref{fig:qualitative} provides a qualitative comparison across three representative tiles.
\method{} concentrates residual errors on complex features, leaving flat and smooth regions with minimal error, illustrating the effect of WCF-guided frequency allocation.
The frequency-domain error maps in the last column show systematic directional streaks for ImplicitTerrain, indicating missing or poorly fitted frequency components.
These artifacts are alleviated in \method{}'s reconstruction, supporting the interpretation that the frequency embedding and WCF masks achieve broader and more accurate spectral coverage.

Since downstream terrain analysis hinges on the shape model's derivative accuracy, we also evaluate gradient fidelity in isolation (Tab.~\ref{tab:shape_grad}).
Gradient matching (GM) reduces GradMAE by $6.4\times$ and worst-case elevation error by $7.7\times$ over the SIREN baseline.
Combining the frequency embedding (FreqEmbed) with GM yields the best results across all metrics, confirming their complementary contributions to elevation and derivative accuracy.

\begin{table}[t]
\centering
\small
\caption{Shape model fitting quality and gradient fidelity. GradMAE measures the mean absolute error of the predicted gradient field against the finite-difference ground truth.}
\label{tab:shape_grad}
\vspace{-0.8em}
\begin{tabular}{lccc}
\toprule
Shape Config & PSNR (dB) $\uparrow$ & GradMAE $\downarrow$ & MaxAE (m) $\downarrow$ \\
\midrule
SIREN                & 55.50 $\pm$ 2.80 & 0.097 & 10.52 \\
+GM           & 67.26 $\pm$ 1.24 & 0.024 &  1.84 \\
+FreqEmbed            & 59.81 $\pm$ 1.93 & 0.086 &  7.93 \\
+FreqEmbed+GM & \textbf{70.54 $\pm$ 1.89} & \textbf{0.015} & \textbf{1.37} \\
\bottomrule
\end{tabular}
\vspace{-0.2cm}
\end{table}

\subsection{Storage Efficiency and Inference}
\label{sec:exp:compression}

Beyond fitting quality, we evaluate \method{} as a terrain data format in terms of storage cost, measured in bits per pixel (bpp, encoded bits divided by the number of DEM grid cells), and query performance.
Table~\ref{tab:compression} compares \method{} against traditional terrain compression methods.
At the recommended 12b+8b operating point (Sec.~\ref{sec:method:compression}), \method{} achieves 1.23\,bpp with only $-0.28$\,dB degradation from the uncompressed model, a $3.2\times$ compression ratio from float32.
The reported bpp is averaged over the 50 tiles and accounts for all stored components %
after quantization and entropy coding, with the per-tile bpp varying primarily through the compressed size of the complexity field.

\begin{figure}[t]
\centering
\includegraphics[width=\columnwidth]{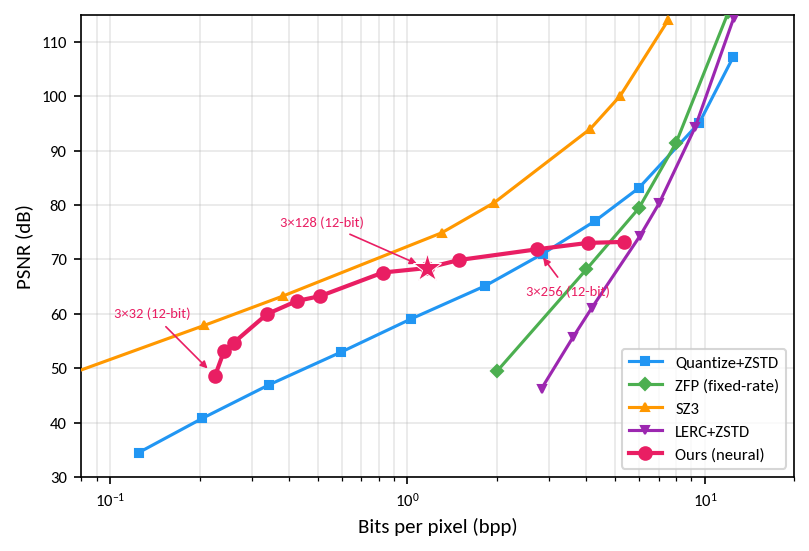}
\caption{Rate-distortion comparison across bitrates. Red points trace our neural format as network width and depth vary under mixed-precision quantization and entropy coding, while traditional codecs are swept over their quality parameters. The {\color{red} $\bigstar$} marker denotes the configuration used throughout the other experiments. The neural format remains competitive across the full bpp range and additionally provides analytical gradients and resolution independence.}
\label{fig:rd_curves}
\end{figure}

Figure~\ref{fig:rd_curves} shows rate-distortion curves obtained by sweeping network width and depth configurations with fixed post-training quantization and entropy coding, compared against traditional codecs swept over their respective quality parameters.
Across the practical bitrate range (0.5--2\,bpp), the proposed neural format outperforms Quantize+ZSTD, ZFP, and LERC+ZSTD, with only SZ3~\cite{liang2022sz3}, an error-bounded compressor particularly suited to smooth scientific data, achieving higher PSNR at matched rates.
The neural format incurs a one-time encoding cost that traditional codecs do not require, and this cost is amortized when the encoded model is queried repeatedly for analysis tasks. It provides capabilities absent from traditional codecs: continuous elevation queries at arbitrary coordinates, analytical gradient computation, and resolution-independent topological analysis.
These capabilities would otherwise require separate processing pipelines on decompressed grid data.

For the model's inference throughput, the compiled model evaluates 77.3\,M coordinate queries per second on an RTX A5000, returning both elevation and gradient in a single pass.
This represents a $4.63\times$ speedup over na\"ive autograd-based gradient computation and is $2.24\times$ faster than na\"ive elevation-only queries without compilation, indicating that the compiled analytical gradient path adds a small overhead to elevation queries.
Floating-point operations (FLOPs), and memory usage are reported in Supp.~\ref{sec:supp:profiling}. Spatial error maps comparing the 12b+8b configuration against classical codecs on a representative tile are provided in Supp.~\ref{sec:supp:compression}

\begin{figure*}[t]
\centering
\includegraphics[width=0.90\textwidth]{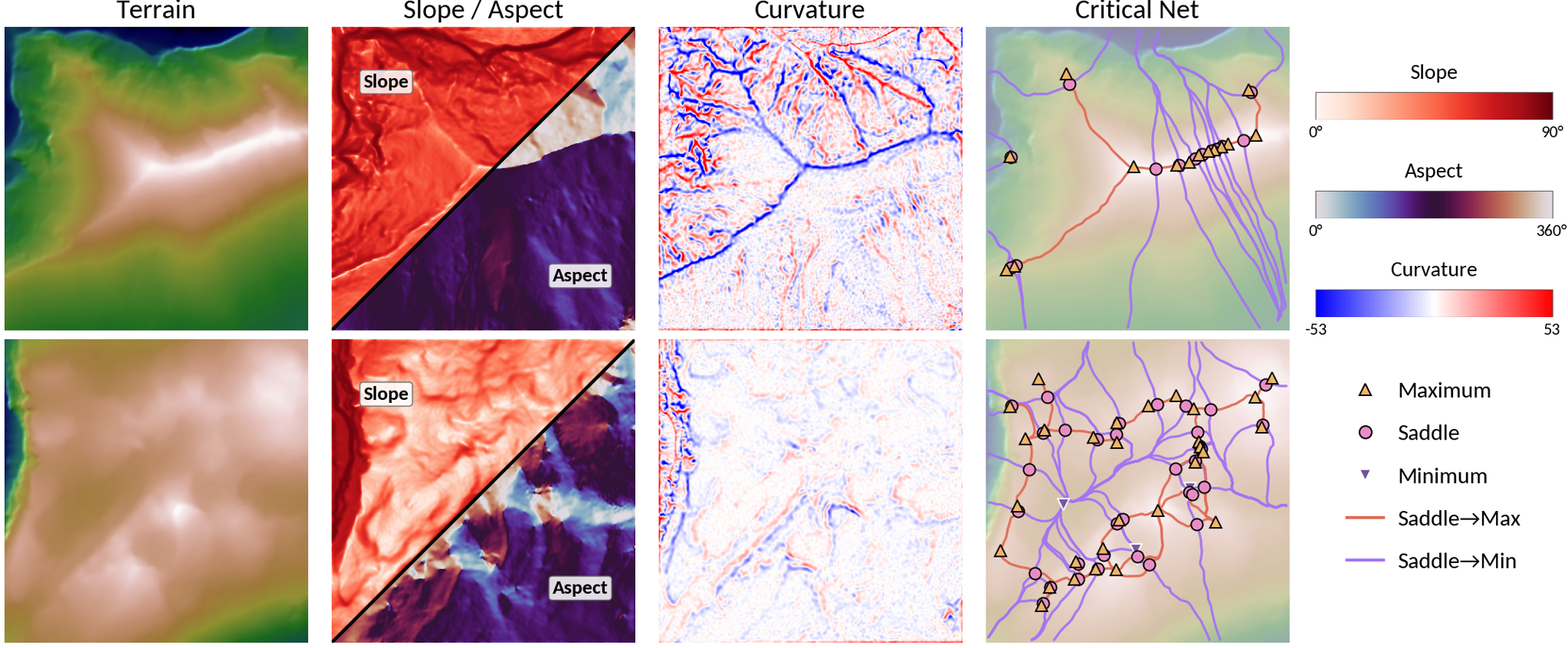}
\caption{Topographic features and topological skeleton derived from \method{}'s shape model $\psis$ via analytical gradients on two representative tiles.
From left to right: elevation (DEM hillshade), combined slope and aspect (split along the anti-diagonal), mean curvature, and the critical network.
Critical points (minima, maxima, saddles) are located at $\|\nabla\psis\| = 0$ and classified via the Hessian eigenvalues. Separatrix lines traced by gradient delineate ridges and valleys.}
\label{fig:topo_features}
\end{figure*}

\begin{table}[t]
\centering
\small
\caption{Storage efficiency comparison.
With mixed-precision quantization and entropy coding, our neural format achieves competitive rate-distortion across the evaluated bitrates.
Classical formats require a full decompression step before any query; our neural terrain format supports queries at arbitrary coordinates with analytical gradient computation, with no decompression required.}
\label{tab:compression}
\vspace{-0.8em}
\begin{tabular}{lrccc}
\toprule
Format & bpp & PSNR (dB) $\uparrow$ & MAE (m) $\downarrow$ \\
\midrule
GeoTIFF~\cite{mahammad2003geotiff}+DEFLATE~\cite{deutsch1996deflate}     & 15.29 & lossless & 0      \\
FPZIP~\cite{lindstrom2006fpzip}               & 11.78 & lossless & 0      \\
LERC~\cite{esri2024lerc}+ZSTD~\cite{collet2018zstd} ($z$=0.1) & 6.06  & 74.35    & 0.045  \\
ZFP~\cite{lindstrom2014zfp} (rate=4)        & 4.00  & 68.19    & 0.057  \\
Quantize+ZSTD~\cite{collet2018zstd} (10b) & 2.86  & 71.08    & 0.092  \\
SZ3~\cite{liang2022sz3} ($\epsilon$=0.05\,m) & 1.96 & \textbf{80.45} & \textbf{0.023}  \\
\midrule
Ours (float32)          & 4.03 & 66.25 & 0.133  \\
Ours (float16)          & 1.77 & 66.25 & 0.133  \\
\textbf{Ours (12b+8b)}  & 1.23 & 65.97 & 0.137  \\
Ours (int8)             & 1.04 & 61.30 & 0.239  \\
\bottomrule
\end{tabular}
\end{table}

\subsection{Ablation Study}
\label{sec:exp:ablation}

\begin{table}[t]
\centering
\small
\caption{Progressive ablation from ImplicitTerrain-128 (IT-128) to \method{} (averaged over 50 tiles). Each row adds one component to the previous.}
\label{tab:ablation}
\vspace{-0.8em}
\begin{tabular}{lccr}
\toprule
Configuration & PSNR (dB) $\uparrow$ & $\Delta$ & Iter (ms) $\downarrow$ \\
\midrule
IT-128 ($\omega_0\!=\!30$)  & 49.74 & --- & 34.5 \\
$+\,\omega_0$ tuning (calibrated baseline) & 62.13 & +12.39 & 34.4 \\
\midrule
$+\,$FreqEmbed + GM         & 63.08 & +0.94  & 34.5 \\
$+\,$WCF masks              & 65.72 & +2.64  & 45.4 \\
$+\,$Adaptive sampling      & \textbf{66.25} & +0.53  & \textbf{20.2} \\
\bottomrule
\end{tabular}
\vspace{-0.2cm}
\end{table}

Table~\ref{tab:ablation} isolates the contribution of each component by progressively adding them to the ImplicitTerrain-128 baseline.
The first row applies an $\omega_0$ correction at the geometry model only; the shape model retains $\omega_0 = 30$ to match the low-frequency spectrum of the Gaussian-smoothed target (Sec.~\ref{sec:exp:setup}).
This targeted adjustment reflects the geometry stage's distinct role of fitting a $1000 \times 1000$ residual dominated by high-frequency geometric detail.
Because ImplicitTerrain~\cite{feng2024implicitterrain} shares $\omega_0 = 30$ across both stages, its geometry stage under-represents the residual content; we therefore regard the resulting $+12.39$\,dB as a baseline calibration specific to the cascaded design rather than a contribution of this work. The subsequent comparisons are made against a properly tuned configuration rather than a weak baseline.
Tab.~\ref{tab:ablation} quantify each module's improvement progressively: the improved shape model with frequency embedding and gradient matching, WCF masks at the geometry model, and adaptive sampling.
Applying the frequency embedding to the geometry model without spatial masks yields limited E2E improvement, and the embedding is retained only as a prerequisite for WCF band gating.
Further analysis of mask placement and per-component overhead is provided in Supp.~\ref{sec:supp:hidden_masks} and~\ref{sec:supp:efficiency}.

\subsection{Terrain Analysis Applications}
\label{sec:exp:analysis}

Beyond fitting accuracy, a key advantage of the neural terrain format is its support for downstream terrain analysis directly from the continuous representation.
Following ImplicitTerrain~\cite{feng2024implicitterrain}, we demonstrate two applications: topographic feature computation and topological analysis.
Both operate on the shape model $\psis$, which captures the smooth manifold structure of the terrain; topological analysis in particular requires a smooth, noise-free surface with accurate gradients, which the shape model provides by design (Tab.~\ref{tab:shape_grad}).
The full elevation reconstruction $\hat{z} = \psis + \psig$ is used for fitting evaluation (Sec.~\ref{sec:exp:fitting}), while the shape model alone serves as the analytical backbone for derivative-based terrain analysis.

Figure~\ref{fig:topo_features} visualizes the topographic quantities and the topological skeleton (\ie, critical net) derived from the shape model $\psis$ on two representative tiles.
Slope, defined as {\small$\arctan(\|\nabla \psis\|)$}, and aspect, defined as $\mathrm{atan2}(\partial \psis/\partial y,\; \partial \psis/\partial x)$, are computed from the first-order gradient $\nabla \psis$.
Mean curvature $\kappa$ can be obtained via the standard differential geometry formula:
\begin{equation}
\kappa = \frac{(1 + \Psi_y^2)\Psi_{xx} - 2\Psi_x \Psi_y \Psi_{xy} + (1 + \Psi_x^2)\Psi_{yy}}{2(1 + \Psi_x^2 + \Psi_y^2)^{3/2}}.
\end{equation}
All quantities require no grid discretization and can be evaluated at arbitrary resolution from the learned model, eliminating the resolution dependence of finite-difference operators.

For topological feature extraction, ImplicitTerrain~\cite{feng2024implicitterrain} demonstrated that continuous SIREN surface models enable Morse-theoretic topological analysis.
Critical points are locations where $\|\nabla \psis\| = 0$; they are located using the Newton-Raphson method~\cite{hildebrand1987introduction} and classified by the eigenvalues of the Hessian: both positive (minimum), both negative (maximum), or mixed signs (saddle).
From each saddle, four separatrix lines are traced by integrating $\nabla \psis$ (ascending toward maxima) and $-\nabla \psis$ (descending toward minima), producing the Morse-Smale complex that delineates the terrain's topological skeleton (ridges, valleys, drainage divides).
\method{}'s improved shape model fitting accuracy and gradient fidelity provide a more accurate continuous field for the same Morse-theoretic analysis supported by~\cite{feng2024implicitterrain}.
The applications in this section are demonstrated qualitatively; the quantitative evidence for derivative-based analysis in this work is the gradient fidelity reported in Tab.~\ref{tab:shape_grad}.

\section{Conclusion}
\label{sec:conclusion}

We presented \method{}, a practical neural data format for terrain that encodes each DEM tile as a compact, spatially-adaptive implicit neural representation. A single stored model provides continuous coordinate queries and analytical gradients, enabling resolution-independent terrain data analysis.
A wavelet complexity field concentrates model capacity in spatially complex regions, while gradient matching improves both elevation reconstruction and gradient fidelity in the smooth shape model.
Together with cascaded SIREN backbones and adaptive sampling, \method{} achieves 66.25\,dB end-to-end PSNR with 124K parameters, improving over ImplicitTerrain by +5.70\,dB while using 3.2$\times$ fewer parameters, 5.8$\times$ less peak GPU memory, and 4.1$\times$ faster training; post-training quantization and entropy encoding of the model weights further compress it to 1.23\,bpp with only a 0.28\,dB PSNR drop.
The resulting format is competitive with several of the evaluated traditional codecs over the tested bitrate range, while offering continuous surface modeling and differentiability as native properties of the representation.

\method{} has the following limitations: the adaptive masks rely on the premise that high-frequency content is spatially sparse~\cite{feng2025sasnet}; each tile is encoded by an independent optimization, so the format targets offline encoding settings where repeated queries amortize this cost; and the evaluation, while morphologically diverse within the swissALTI3D dataset, does not establish generalization to other landform distributions such as low-relief plains, coastal, or volcanic terrain.
Future work includes meta-learning-based encoding~\cite{dupont2022coin++} to amortize the per-tile optimization cost, integration with existing geospatial toolchains, and evaluation on a broader range of landforms and on other geospatial surfaces with smooth derivatives such as bathymetry and planetary DEMs.

\begin{acks}
This work has been partially supported by the US National Science Foundation under grant numbers IIS-1910766 and IIS-2114451.
\end{acks}

\bibliographystyle{ACM-Reference-Format}
\bibliography{main}

\clearpage
\appendix

\setcounter{figure}{0}
\setcounter{table}{0}

\section{Dataset Curation}
\label{sec:supp:dataset}

We evaluate on 50 terrain tiles selected from 33{,}399 candidates from the swissALTI3D DEM~\cite{swissALTI3D} via a two-pass pipeline.
Each raw tile covers $2000\times 2000$ pixels at 0.5\,m resolution, which we downsample to $1000{\times}1000$ (1\,m) for training.
\emph{Pass~1 (quality)} computes texture and frequency descriptors on $256{\times}256$ versions ($\sim$17\,ms/tile) and rejects tiles with missing data, elevation range below 50\,m, near-zero surface variation, or more than 70\% low-gradient pixels (32{,}937 of 33{,}399 accepted).
\emph{Pass~2 (diversity)} computes a 10-dimensional descriptor (elevation range, normalized std, mean and 95th-percentile gradient, low-gradient ratio, high-frequency spectral energy, 90th/95th-percentile frequencies, spectral slope, Haar-wavelet complexity), z-score normalizes the features, and applies greedy farthest-point sampling (seed\,=\,42).
The 50 selected tiles span gentle valleys (21 tiles, $<$200\,m elevation range), hilly foothills (19 tiles, 200--500\,m), and alpine terrain (10 tiles, $>$500\,m, up to 1{,}667\,m), with absolute elevations 101--1{,}668\,m (mean 339\,m), wavelet complexity 0.33--0.86, and low-gradient ratio 0.06--0.70.
For preprocessing, each tile is bilinearly downsampled to $1000\times 1000$ as the geometry-model ground truth. The shape-model target additionally normalizes to $[0,1]$, Gaussian-smooths, and downsamples to $ 500 \times 500$.
Each tile is independently fitted by the INR with its own weights.%

\section{Hidden Layer Mask Ablation}
\label{sec:supp:hidden_masks}

The original SASNet~\cite{feng2025sasnet} applies spatial masks to both the input layer and all hidden layers (8 neuron groups per hidden layer), a design motivated by RGB image fitting, where repeated textures benefit from location-conditioned gating throughout the network.
To test whether this carries over to terrain, we run SASNet with a hash-grid encoder ($\omega_0{=}150$, 2{,}000 iterations) on 5 tiles that span the dataset's complexity range.
Enabling hidden-layer masks \emph{degrades} geometry-model PSNR by 0.58\,dB on average (40.17\,dB without vs.\ 39.59\,dB with), with per-tile drops ranging from 0.13 to 1.05\,dB.
The hidden-layer mask decoders add only 1{,}176 parameters ($+1.3\%$), so the degradation is not driven by reduced capacity but more likely by group-level gating in the hidden SIREN layers, limiting cross-group feature interaction, which is unhelpful for smooth, spatially non-repetitive terrain residuals.
In this ablation, input-layer spatial adaptivity alone (selecting which frequency bands are active at each location) performs best, motivating our design choice to apply WCF masks only at the first layer.

\section{Efficiency, Training, and Numerical Analysis}
\label{sec:supp:efficiency}

\vspace{0.2em}\noindent\textbf{Gradient matching (GM) overhead.}\label{sec:supp:gm_cost}
GM computes $\partial f / \partial \mathbf{x}$ via PyTorch autograd on a 10K-pixel random subsample of the $500{\times}500$ shape grid drawn each iteration, and matches against finite-difference ground-truth gradients.
On the shape model, it adds $+7.8\%$ peak memory (1{,}243 to 1{,}340\,MB) and $+11.8\%$ per-iteration time (17.96 to 20.07\,ms), nearly tile-independent ($\pm 0.04$\,ms std) and dominated by autograd overhead.
Since GM runs only during the 3{,}000-iteration shape phase, end-to-end pipeline overhead is $+5.0\%$, and the 10K subsample reduces measured GM FLOPs by $16.5\times$ vs.\ full-grid gradient computation, with no observed change in our metrics.

\vspace{0.2em}\noindent\textbf{WCF and adaptive sampling (AS).}
For the geometry model on a $1000{\times}1000$ tile under \texttt{\small torch.compile}+AMP, the WCF decoder adds $+15.6$\,ms per iteration ($+50.8\%$ over a 30.6\,ms SIREN baseline), 14\% less than the hash-grid alternative ($+59.2\%$), and uses 14.6K fewer mask parameters (24K vs.\ 39K).
With 25\% complexity-guided AS, the full WCF+AS pipeline runs at 20.2\,ms per iteration ($-34.1\%$ vs.\ SIREN baseline at matched iterations): the $4\times$ pixel reduction dominates the decoder cost.
Peak GPU memory drops from 5.9\,GB to 1.8\,GB ($3.2\times$ reduction).

\begin{figure}[t]
\centering
\includegraphics[width=\columnwidth]{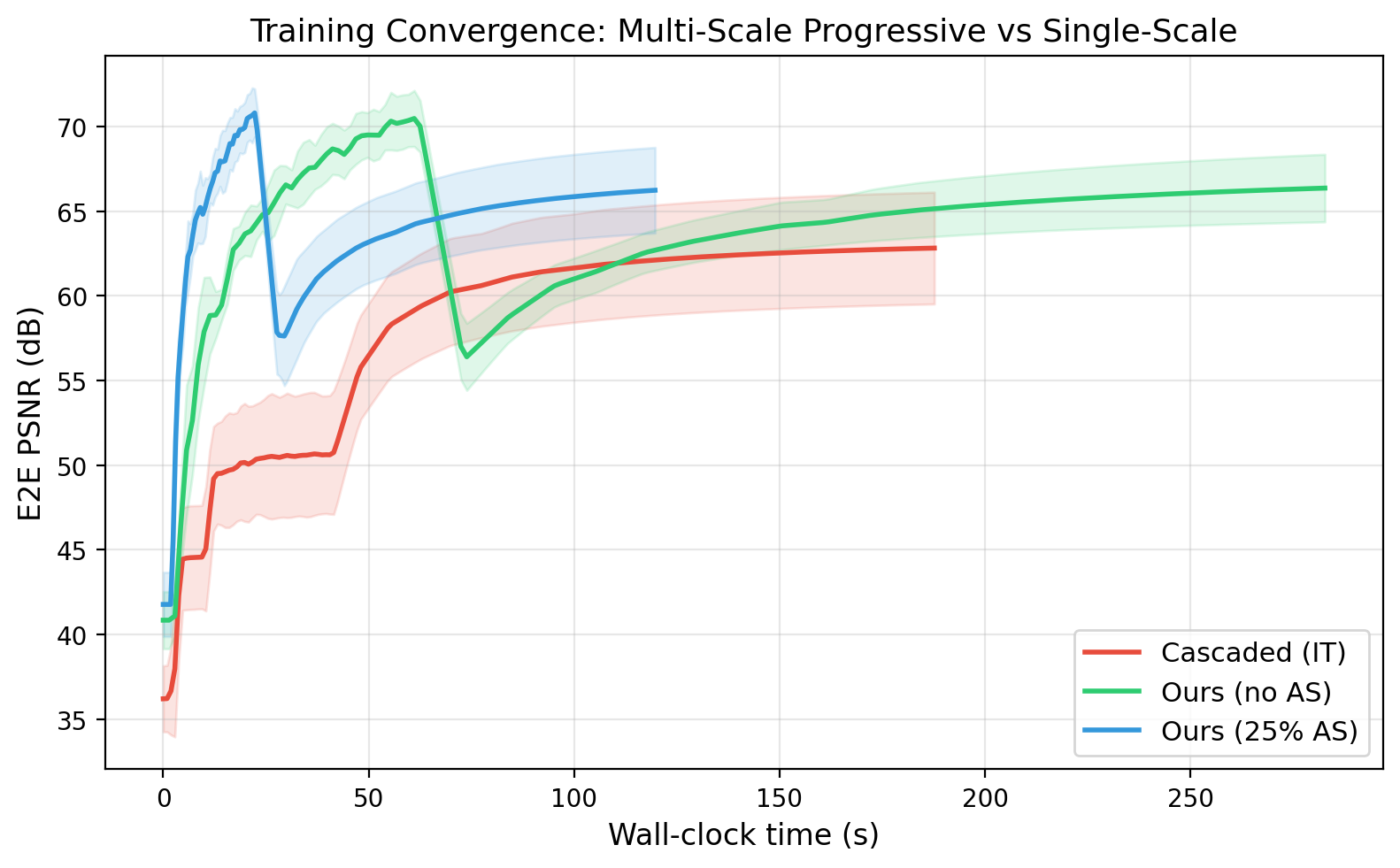}
\vspace{0em}
\caption{Convergence on 5 tiles: our single-scale WCF pipeline with 25\% AS (green) reaches higher E2E PSNR faster than cascaded pyramid baseline~\cite{feng2024implicitterrain} (blue).}
\label{fig:supp:convergence}
\end{figure}

\vspace{0.2em}\noindent\textbf{Multi-scale vs.\ single-scale training.}\label{sec:supp:multiscale}
ImplicitTerrain~\cite{feng2024implicitterrain} trains its shape model using Gaussian pyramid progressive training (coarse-to-fine resolutions).
As shown in Fig~\ref{fig:supp:convergence}, on 5 randomly selected tiles and under matched iteration budgets (3{,}000 shape + 2{,}000 geometry) with the same $3{\times}128$ backbone and GM, our single-scale pipeline with WCF reaches $66.39 \pm 2.00$\,dB E2E PSNR vs.\ $62.83 \pm 3.30$\,dB for the cascaded pyramid, and with 25\% AS reaches $66.25 \pm 2.53$\,dB ($+3.42$\,dB over the pyramid) at $1.57\times$ faster wall-clock time (120.6 vs.\ 189.4\,s, full-precision protocol shared with ImplicitTerrain).

\vspace{0.2em}\noindent\textbf{Shape-model sampling.}\label{sec:supp:shape_as}
Because the shape target is Gaussian-smoothed (low-frequency), uniform subsampling reduces cost with negligible quality change.
On 10 tiles, both 50\% and 25\% uniform subsampling lose only $-0.01$\,dB PSNR with unchanged GradMAE to four decimals, providing $1.27\times$ and $1.84\times$ speedups.
At 25\%, $\sim$62.5K pixels still exceed the model's 50K parameters, and the smooth signal is spatially redundant. The geometry model, by contrast, requires complexity-guided sampling because its residual is spatially non-uniform.
We adopt 25\% uniform subsampling as the default for shape-model training.

\begin{figure*}[t]
\centering
\includegraphics[width=\textwidth]{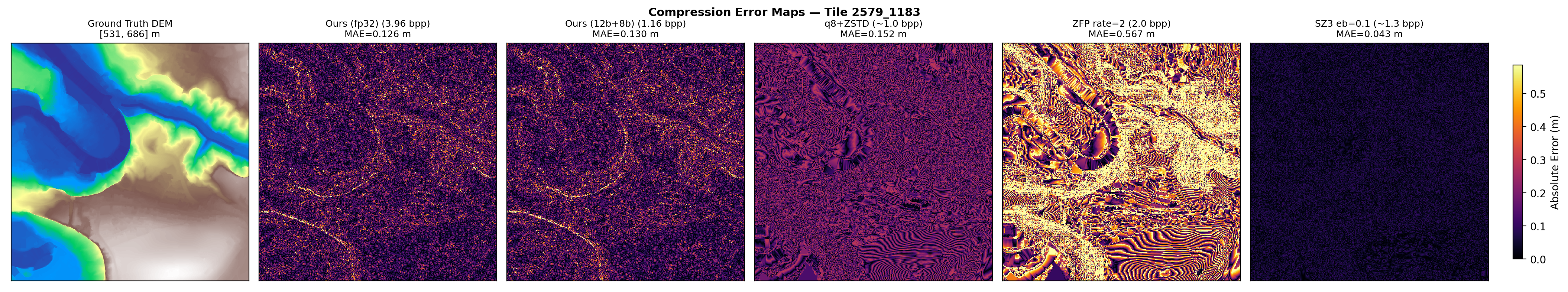}
\vspace{0em}
\caption{Absolute-error maps on a tile with 155\,m relief (shared colorscale in m).}
\label{fig:supp:error_maps}
\end{figure*}

\vspace{0.2em}\noindent\textbf{Profiling.}\label{sec:supp:profiling}
All 128-width methods share an identical 50K-parameter SIREN backbone per stage, differing only in auxiliary modules, while the cascaded baseline uses a $3{\times}256$ backbone (198K/stage, no auxiliary).
SASNet adds a 39K hash-grid mask branch in the geometry stage, whereas our WCF decoder uses 24K, saving 14.6K auxiliary parameters while achieving higher accuracy.
Per-query forward FLOPs are 99K ($1.00\times$) for SIREN-128, 101K ($1.02\times$) for SASNet hash-grid, 147K ($1.49\times$) for WCF, and 395K ($3.98\times$) for the $3{\times}256$ baseline.
The $4\times$ pixel reduction from AS yields a net $2.25\times$ training speedup despite the higher per-query cost.
Per-stage peak GPU memory (shape/geometry) is 1{,}340/1{,}740\,MB for ours, vs.\ 2{,}464/9{,}800 for the $3{\times}256$ baseline and 1{,}241/5{,}436 for SASNet, and AS makes our geometry stage comparable to the shape stage.
The deployed artifact stores the 100K-parameter SIREN backbone plus the 24K WCF decoder, scales, learned thresholds, normalization statistics, and a 4-bit, entropy-coded low-resolution complexity field, all counted toward the reported 1.23\,bpp (averaged across 50 tiles, since the compressed complexity field varies by tile). At inference, the normalized complexity field $\hat{c}$ at each query coordinate is obtained by bilinear interpolation of the decoded low-resolution field, and the $K{=}4$ band masks are then produced via the learned thresholds, matching the pathway used during training.

\vspace{0.2em}\noindent\textbf{Inference and numerical notes.}
On an RTX A5000 (50 warmup + 200 timed iterations), the compiled analytical-gradient path reaches 77.3\,Mq/s at 1M-query batch and saturates beyond 100K queries, a $2.24\times$ speedup over na\"ive elevation-only (34.6\,Mq/s), so the optimized path returns elevation \emph{and} gradient faster than the unoptimized path returns elevation alone, due to \texttt{\small torch.compile} fusing sin/cos/matmul kernels and AMP halving memory bandwidth.
End-to-end training takes 55\,s/tile with \texttt{\small torch.compile}+AMP, covering the cascaded shape phase (3{,}000 iterations) and geometry phase (2{,}000 iterations at WCF + 25\% AS).
Native float16 SIREN training produces NaN on all tested tiles because the $\sin(\omega_0 \mathbf{Wx}{+}\mathbf{b})$ activation accumulates rounding error through 3 hidden layers that exceeds the signal amplitude. AMP (forward in float16, master weights in float32, with gradient scaling) matches float32 to within $0.01$\,dB at $1.94\times$ speedup.

\section{Compression and Rate-distortion Analysis}
\label{sec:supp:compression}

\vspace{0.2em}\noindent\textbf{Per-layer quantization sensitivity.}\label{sec:supp:quant_sensitivity}
We sweep per-layer PSNR drop when quantizing each layer group individually with all others at float32, using weight-only symmetric uniform quantization with per-output-channel weight scales and 16-bit biases. Input-layer weights are regenerable from the frequency table and skipped in the deployed format, with sensitivity values reported only for reference.
Table~\ref{tab:supp:quant} reports $\Delta$\,PSNR for both stages.
Two patterns emerge. First, sensitivity decreases from input to output layers, since input-layer perturbations are amplified by $\omega_0$ through subsequent sine activations. Second, the shape model ($\omega_0{=}30$) is substantially more sensitive than the geometry model ($\omega_0{=}150$) at the same bit-width, because the geometry model fits simpler residual signals with smaller weight magnitudes. The WCF decoder is nearly insensitive since it produces binary masks robust to perturbations.
At 12-bit, all shape-model layers stay within $-0.83$\,dB and all geometry-model layers within $-0.04$\,dB, motivating the mixed-precision configuration: 12-bit for the shape model (preserving gradient accuracy) and 8-bit for the geometry model and WCF decoder, yielding 1.23\,bpp at only $-0.28$\,dB E2E across 50 tiles.

\begin{table}[t]
\centering
\small
\setlength{\tabcolsep}{4pt}
\caption{Per-layer quantization sensitivity: $\Delta$\,PSNR (dB) when quantizing one layer group with others at float32. Shape baseline 68.37\,dB, geometry baseline 44.52\,dB.}
\label{tab:supp:quant}\vspace{-0.4cm}
\begin{tabular}{llrrrr}
\toprule
Stage & Layer & 16b & 12b & 10b & 8b \\
\midrule
\multirow{5}{*}{Shape}
 & Input    & $-0.00$ & $-0.83$ & $-5.69$ & $-17.69$ \\
 & Hidden 0 & $+0.00$ & $-0.03$ & $-1.47$ & $-8.60$ \\
 & Hidden 1 & $-0.00$ & $-0.04$ & $-0.83$ & $-6.36$ \\
 & Hidden 2 & $+0.00$ & $+0.00$ & $-0.09$ & $-1.61$ \\
 & Output   & $-0.00$ & $-0.00$ & $-0.19$ & $-1.78$ \\
\midrule
\multirow{6}{*}{Geometry}
 & Input    & $-0.00$ & $-0.04$ & $-0.52$ & $-5.26$ \\
 & Hidden 0 & $-0.00$ & $-0.01$ & $-0.16$ & $-2.05$ \\
 & Hidden 1 & $-0.00$ & $-0.00$ & $-0.05$ & $-0.87$ \\
 & Hidden 2 & $-0.00$ & $-0.00$ & $-0.01$ & $-0.11$ \\
 & Output   & $-0.00$ & $-0.00$ & $-0.00$ & $-0.00$ \\
 & WCF dec. & $+0.00$ & $-0.00$ & $-0.00$ & $-0.03$ \\
\bottomrule
\end{tabular}
\vspace{-0.1cm}
\end{table}

\vspace{0.2em}\noindent\textbf{QAT vs.\ PTQ.}\label{sec:supp:qat}
On 5 tiles, QAT with a straight-through estimator gives negligible improvement at int8 ($+0.02$\,dB, 60.87 vs.\ 60.85) while substantially degrading float32 quality ($-4.52$\,dB, 61.29 vs.\ 65.81): fake-quantization noise during training appears to hinder the model from learning fine spatial details.
At our 12b+8b target operating point, QAT also underperforms PTQ (61.19 vs.\ 65.54), because 12-bit PTQ already incurs only small loss on the standard-trained model.
We hypothesize that SIREN's frequency amplification $\sin(\omega_0 \mathbf{Wx}{+}\mathbf{b})$, which magnifies weight perturbations by $\omega_0$, creates a precision floor at 8-bit that is difficult to overcome through training-time optimization alone.
QAT narrows the float32-to-int8 gap (from $-4.96$ to $-0.42$\,dB) by lowering the float32 ceiling rather than raising the int8 floor.
We therefore adopt PTQ for the neural terrain format, as QAT adds training complexity without measurable benefit at the target bit-widths.

\vspace{0.2em}\noindent\textbf{Compression error maps.}
Fig.~\ref{fig:supp:error_maps} visualizes spatial error distributions on a representative steep-foothills tile.
On this illustrative tile, our 12b+8b configuration reaches MAE/MaxAE of 0.130\,m / 6.02\,m, vs.\ q8+ZSTD at 0.152\,m / 0.31\,m, ZFP rate=2 (2.0\,bpp) at 0.567\,m / 49.14\,m, and SZ3 ($\epsilon{=}0.1$, $\sim$1.3\,bpp) at 0.043\,m / 0.10\,m.
Neural errors concentrate at complex features (ridgelines, cliff edges, steep slopes) and stay small in flat regions, so worst cases occur at spatially predictable locations rather than scattered across smooth regions.

\end{document}